\documentclass{article}
\usepackage{aaai19}

\usepackage{times}
\usepackage{xcolor}
\usepackage{soul}
\usepackage{url}
\usepackage[utf8]{inputenc}
\usepackage[small]{caption}
\usepackage{booktabs}
\usepackage{mathrsfs}
\usepackage{amssymb}
\usepackage[scr]{rsfso}
\usepackage[mathscr]{euscript}
\usepackage{amssymb,amsmath}
\usepackage{bm}
\usepackage{graphicx}
\usepackage{caption}
\usepackage{subfigure}
\usepackage[T1]{fontenc}
\usepackage{cleveref}
\usepackage{color}
\usepackage{bbm}
\usepackage{booktabs}
\usepackage{float}
\usepackage{graphicx,dblfloatfix}
\usepackage{multirow}
\newcommand{\depth}{\mathit{Depth}}
\newcommand{\perm}{\mathit{Perm}}

\newcommand{\ddpthall}[0]{Differentiable Depth}
\let\oldmarginpar\marginpar
\renewcommand\marginpar[1]{\-\oldmarginpar[\raggedleft\footnotesize\color{red} #1]{\raggedright\footnotesize\color{red} #1}}

\newcommand{\zap}[1]{}
\newcommand{\real}[0]{\mathbb{R}}
\newcommand{\mM}[0]{\mathcal{M}}
\newcommand{\mU}[0]{\mathcal{U}}
\newcommand{\mO}[0]{\mathcal{O}}
\newcommand{\NNsname}[0]{$\text{MPNN}_3$}
\newcommand{\colvec}[2]{\begin{small}\left[\begin{matrix}#1 \\ #2\end{matrix}\right]\end{small}}
\title{Fast OBDD Reordering using Neural Message Passing on Hypergraph}

 \author{
Feifan Xu$^1$,
Fei He$^1$,
Enze Xie$^2$,
Liang Li$^1$
\\
$^1$ Tsinghua University \\
$^2$ Tongji University  \\
 }

\begin{document}
  \maketitle

%
%
\begin{abstract}
  Ordered binary decision diagrams (OBDDs) are an efficient data structure
    for representing and manipulating Boolean formulas.
  With respect to different variable orders, the OBDDs'
  sizes may vary from linear to exponential
  in the number of the Boolean variables.
  Finding the optimal variable order has been proved a NP-complete problem.
  Many heuristics have been proposed to find a near-optimal solution of
  this problem.
In this paper, we propose a neural network-based method
to predict near-optimal variable orders for unknown formulas.
Viewing these formulas as hypergraphs,
and lifting the message passing neural network into 3-hypergraph (\NNsname),
we are able to learn the patterns of Boolean formula.
Compared to the traditional methods, our method can find a near-the-best
solution with an extremely shorter time, 
even for some hard examples.
To the best of our knowledge, this is the first work on applying neural network to OBDD reordering.

\end{abstract}

  \section{Introduction}

Boolean functions are functions that take Boolean variables as arguments and return
Boolean values.
They were widely accepted as the modeling formalism for design, verification and synthesis of digital
computers\cite{crama2011boolean}.
More real-world problems, including cryptography, social choice theory, etc.,
can be formulated using Boolean functions.

Ordered binary decision diagrams (OBDDs)~\cite{Bryant:ObddIntro} are a standard
  data structure for representing and manipulating Boolean formulas.
  They are compact to store and efficient to operate. More importantly, they
  provide a canonical representation of Boolean functions. Given any two
  logically equivalent Boolean functions, their OBDDs are isomorphic.

A central problem in the application of OBDDs is to find a proper decision order
of Boolean variables.
With respect to different decision order, the OBDDs' sizes may vary from linear to exponential
in the number of the variables~\cite{bryant1986graph}.
Obviously, we prefer the decision order that minimizes the OBDD's size.

However, the problem of finding the optimal order of an OBDD
is NP-complete~\cite{Bollig:NPprove}.
Many heuristics have been proposed to find a near-optimal solution of this problem.
However, all existing techniques do not achieve a good balance between efficiency
and effectiveness.
Methods that can significantly reduce the OBDD's size always take a long time.
Methods that  take advantages in speed always can't achieve significant results.
Although the problem is NP-complete, the Boolean formulas generated from real world (e.g. circuits, programs, etc.) \emph{do have some patterns}.
If we can utilize these patterns, it is possible to develop a technique that
is both efficient and effective.

NNs have been applied to many areas, including computer
vision, natural language processing, recommendation systems, etc.
Surprisingly, there is no work for applying deep learning to the OBDD
reordering problem.
One possible reason is
 most NN frameworks are not suitable for learning on boolean formulas.
 For example,
Recurrent Neural Networks (RNNs) \cite{RNNs} family can handle sequences learning problem.
To apply RNNs, we must serialize Boolean formulas (usually in \emph{3-conjunctive normal form} (3-CNF))  into sequences.
However, Boolean formula have rich invariances that such a sequential model would ignore,
such as the \emph{permutation invariance} of clauses~\cite{learnSAT}.
For example,  $x_1\land(x_2\lor x_3)$ and $(x_2\lor x_3)\land x_1$ are
syntactically different but semantically equivalent.
The sequential model will \emph{ignore} the permutation invariance, take them as different input.

CNFs can be viewed as hypergraphs~\cite{CNFasHgraph2}.
If we can directly apply deep learning on hypergraph, the semantics of Boolean formula will not be wrecked.
The message passing neural network (MPNN) framework~\cite{MPNN} is a powerful deep
learning technique for graphs.
However, it cannot be utilized directly on hypergraph since the message function is defined on ordinary edges.
In this paper, we lift the MPNN framework to  3-hypergraphs (\NNsname),   define a message function on hyperedges.
We implement the framework in gated graph neural network(GGNN).
In ordinary graphs, GGNN models edges  by  square matrices, then
 uses matrix multiplication to \emph{generate message}.
We use non-square matrices to model hyperedges so that message can be generated and passed on hyperedges.

Compared with the existing techniques, \NNsname~can give a near-optimal solution in an extremely short time,
even for some hard examples that are unsolvable with the existing techniques.
The main technical contributions of this paper are summarized as follows:
\begin{itemize}
    \item We view OBDD reordering  as deep learning  on 3-hypergraph.
      To the best of our knowledge, this is the first  neural network-based approach for OBDD reordering.
    \item Following the main idea of message passing, we lift the MPNN to
        3-hypergraph. 
    \item
      Experimental results show that our approach can find a near-optimal
      order in an extremely short time.
\end{itemize}

\zap{
The remainder of this paper is organized as follows:
  in Section 2, we recall some basic notations of Boolean functions and OBDDs;
  in Section 3, we reformulate the OBDD reordering as a neural network problem;
  Section 4 proposes a successive neural network to solve the problem;
  Section 5 reports experimental results;
  and Section 6 concludes the paper. 
  }

\subsection{Related Work}
{Many OBDD reordering algorithms} have been proposed in the literatures.
~\cite{fujita1991variable} and ~\cite{ishiura1991minimazation} propose the
  \emph{window permutation algorithm} by exchanging a variable
    with its neighbor in the ordering.
~\cite{rudell1993dynamic} proposes the \emph{sifting algorithm} which finds the
optimum position of a variable by repeatedly move it forward or backward.
~\cite{linear}  applies \emph{linear transformations} to minimize OBDD's size.
~\cite{bollig1995simulated} applies \emph{simulated annealing}
  to find a near-optimal order.
  Instead of using swap or exchange operation,
  this method defines a \emph{jump} operation. 
~\cite{drechsler1996genetic} uses the \emph{genetic algorithm}
  to optimize the OBDD's size.
Among all these heuristics, the genetic algorithm and the simulated annealing
algorithm attain the best results but are also the most time consuming.
\cite{JAIRLearnBDD} uses decision tree to learn  variable pair permutation which is more likely to lead to a good order.
In contrast, our NN approach can directly produce the total order of all variable, not pairwise.

  There exist some works on applying neural networks to the OBDD-related topics.
  \cite{Beg:complexity} applies the feed-forward and recurrent neural network
    to predict the OBDD's size for a given Boolean function.
  ~\cite{Bartlett:chooseH2002} studies the problem of converting fault trees
  to OBDDs.
  They propose a neural network approach for selecting one among several
  existing heuristics to construct the OBDD.
  Their approach is essentially a heuristic selection mechanism, and heavily
  depends on the available heuristics.
  In contrast, our approach can directly produce an OBDD variable order.
\cite{learnSAT} use  message passing neural networks to learn to solve SAT problems.
They convert CNFs into graphs, view both literal and clause as node.
In our method clause is treated as hyperedges.



  \section{Preliminaries}
\subsection{Boolean Functions}
Let $\mathbb{B}$ be the \emph{Boolean domain}. 
Let $X = \{x_1, x_2, \cdots, x_n\}$ be a set of \emph{Boolean variables} over
$\mathbb{B}$.
A truth assignment decides a truth value (either 0 or 1) for each
variable in $X$.
A \emph{Boolean function} $f$ over $X$ is a function that takes $x_1, x_2, \cdots,
x_n$ as the arguments and returns either 0 or 1.
The size $|X|$ is called the \emph{arity} of $f$.
The formula $f(X)$ is also called a Boolean formula.
A truth assignment satisfies $f$ iff taking this truth assignment as arguments,
$f$ returns 1.

Let $x$ be a Boolean variable, a \emph{literal} of $x$ is either its positive
form (i.e. $x$) or its negative form (i.e. $\bar{x}$).
A  \emph{clause} is a   disjunction of several literals.
A  \emph{conjunctive normal form (CNF)} is a conjunction of several clauses.
For example, $(x_0 \vee x_1) \wedge (x_0 \vee \bar{x}_2)$ is a CNF.
For simplicity, we often ignore the $\wedge$ operators in a formula.


A 3-CNF is a CNF where all clauses have three or less literals.
Any Boolean formula can be transformed into an equisatisfiable 3-CNF formula~\cite{tseitin1968complexity}.
In the remainder of this paper, we assume all Boolean formulas are in 3-CNF.
\zap{
Most of the popular boolean function file formats, are based on CNF or DNF,
such as PLA and DIMACS, which is also the input file types to build OBDDs.
}

\subsection{Binary Decision Diagrams}

A \emph{binary decision diagram (BDD)} is a rooted, directed acyclic
graph $G = ( V, E )$ with a node set $V$ and an edge set $E$.
Two types of nodes are contained in $V$, i.e., the terminal nodes and the nonterminal nodes.
A terminal node has no outgoing edge, and is labelled with either 0 or 1.  
A nonterminal node $v$ is labelled with a variable $x$ (called the
\emph{decision variable} at this node), and has two successors,
$low(v)$ and $high(v)$, where
$low$ and $high$ indicate the decided values of $x$ being $0$ and
$1$, respectively.
\Cref{fig:ObddPic} shows two BDDs, where the edges
to $low(v)$ and $high(v)$ are marked as dotted
and solid lines, respectively.

Let $G_f$ be a BDD of $f$. Let $\varphi$ be a truth assignment to $X$.
We can easily decide if $\varphi$ satisfies $f$ by traversing $G_f$ from its root to
one of its terminal nodes. Let $v$ be the current node.
If the variable $x$ labelled by $v$ is assigned 0 in $\varphi$, the next node on
the path is $low(v)$;
otherwise, if the variable $x$ is assigned 1, the next node is
$high(v)$.
The value that labels the final reached terminal node gives the value of the function.
Taking the left BDD in \Cref{fig:ObddPic} as an example, with a truth assignment of
$x_0=0, x_1=0, x_2=0, x_3=0, x_4=0$ and $x_5=0$, the value of $f$ can
be quickly decided to be 0 by traversing the graph.

\zap{
A BDD with a root node $v$ represents a Boolean function $f_v$ that:
  \begin{itemize}
  \item If $v$ is terminal, $f_v=value(v)$
  \item If $v$ is nonterminal and $i=index(v)$, then
    \begin{small}
      $$
        \begin{aligned}
          f_v(x_1, ..., x_n)=&{\bar x}_i\land f_{\,low(v)~}(x_1, ..., x_{i-1}, x_{i+1}, ..., x_n)\lor\\
          &x_i\land f_{high(v)}(x_1, ..., x_{i-1}, x_{i+1}, ..., x_n)
        \end{aligned}
      $$
    \end{small}
  \end{itemize}
}

Let $\prec$ be a total order on $X$.
An ordered BDD (OBDD) with respect to $\prec$ is a BDD such that the decision order of variables on all
paths of this OBDD follow $\prec$.
A reduce algorithm~\cite{bryant1986graph} can be repeatedly applied,
to eliminate the possible redundancies in an OBDD.
  The resulting structure is called a reduced OBDD.
  Note that both BDDs in \Cref{fig:ObddPic} are reduced OBDDs.

Reduced OBDD is a canonical representation for Boolean functions. Given any
two logically equivalent Boolean functions, their reduced OBDDs with respect
to a variable order are isomorphic~\cite{bryant1986graph}.
In this paper, we assume all OBDDs are reduced OBDDs.

\subsection{Variable Reordering Problem}

\begin{figure}[tb]
  \centering
  \includegraphics[width=0.4\textwidth]{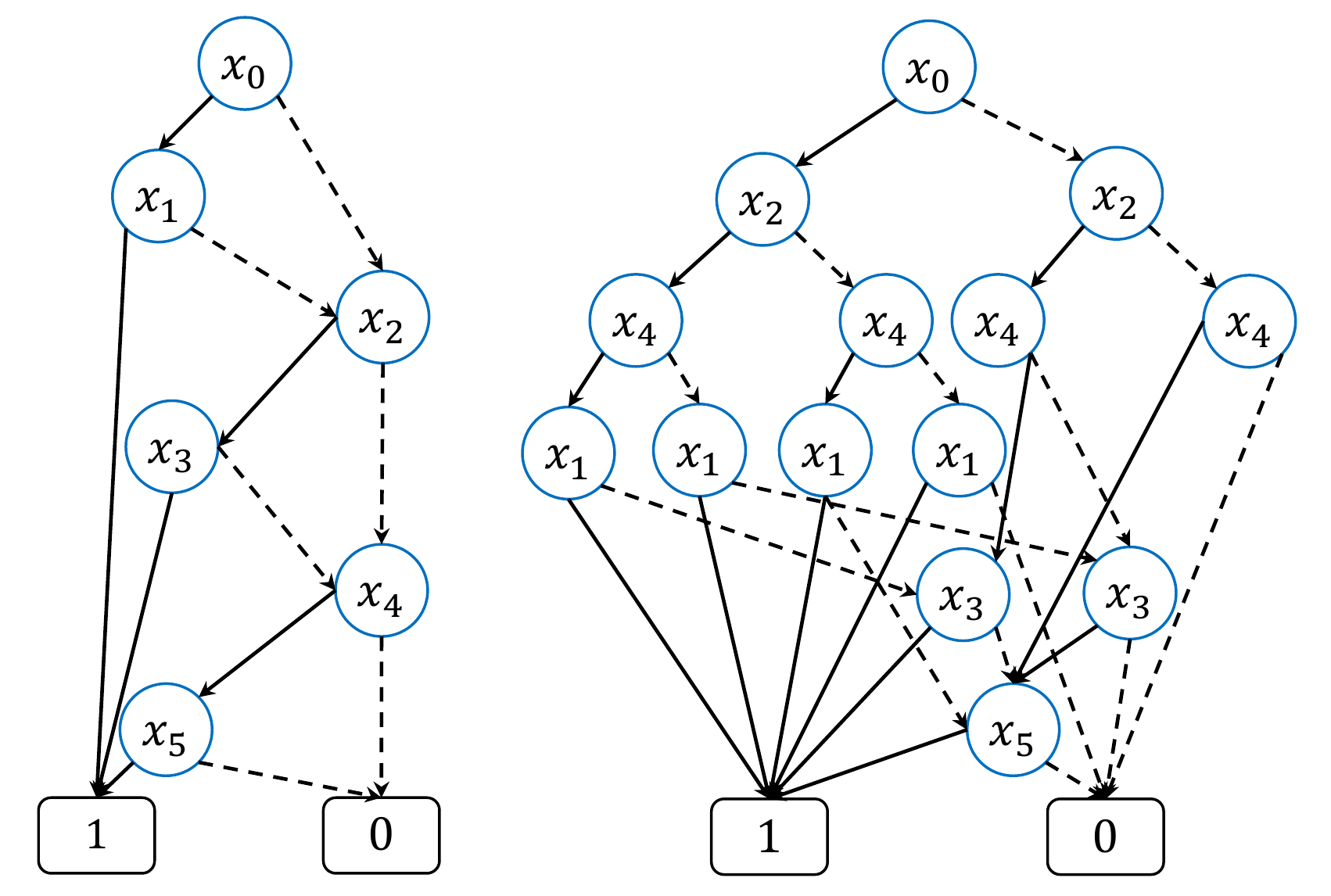}
  \caption{OBDDs of $x_0 x_1 \lor x_2 x_3 \lor x_4 x_5$}
  \label{fig:ObddPic}
\end{figure}

Given an OBDD $G$, we denote the size of $G$ by $|G|$, i.e., the number of nodes in $G$.
The size of an OBDD is highly sensitive to its variable order.
Consider a Boolean function $f=x_0x_1\lor x_2x_3\lor x_4x_5$.
  If we choose a variable order of
    $x_0\prec x_1 \prec x_2 \prec x_3 \prec x_4 \prec x_5$,
    the OBDD's size is $8$ (the left OBDD in \Cref{fig:ObddPic}).
  In contrast, if we choose another variable order of
    $x_0\prec x_2 \prec x_4 \prec x_1 \prec x_3 \prec x_5$,
    the corresponding OBDD's size is 16 (the right OBDD in \Cref{fig:ObddPic}).

In general, for a Boolean function of the
  form $f=x_0x_1\lor x_2x_3\lor ... \lor x_{n-2}x_{n-1}$,
  with the variable order $x_0\prec x_1\prec \cdots\prec x_{n-1}$, its OBDD's size is $n+2$;
while with the variable order $x_0\prec x_2 \prec \cdots\prec x_{2k-2} \prec
x_1\prec x_3\prec \cdots\prec x_{2k-1}$ (assume $n=2k$), its OBDD's size
becomes $2^{\frac{n}{2}+1}$.
In other words, with repsect to different variable orders, the OBDD's size of
a Boolean function may vary from linear to exponential in the number of
variables~\cite{bryant1986graph}.

The \emph{OBDD reordering problem} is to find an optimal variable order for a
given Boolean function, such that its OBDD's size is minimal. This problem has
been proved NP-complete~\cite{Bollig:NPprove}.
Many heuristics~\cite{fujita1991variable,ishiura1991minimazation,rudell1993dynamic,drechsler1996genetic,bollig1995simulated} have been proposed to find a near-optimal solution for
this problem.
Among all these heuristics, the genetic
algorithm and simulated annealing algorithm often attain the
best results~\cite{drechsler1996genetic,bollig1995simulated}.
However, both these two algorithms are quite time consuming.
We seek for a reordering algorithm that can not only find a near-optimal
solution, but also be time efficient.

\subsection{Graph and Hypergraph}

A \emph{graph} $G=(V,E)$ is defined by a set of \emph{vertices} (also called \emph{nodes}) $V=\{1,2,\cdots,|V|\}$ and a set of edge $E\subset V\times V$ which defines the relation between nodes.
A \emph{hypergraph} $H=(V,\hat{E})$ is a generalization of a graph in which an
edge can connect more than two vertices,
and thus $\hat{E} \subset \mathcal{P}(V)$, where $\mathcal{P}$ means power set.
The  $k$-uniform hypergraph is a hypergraph such that all its hyperedges have
exactly $k$ nodes.
We use \emph{$k$-hypergraph} to represent the set of all $k$-uniform hypergraph.
 In this paper, we consider only 3-hyperedges i.e. $\hat{E} \subset V^3$.

Let $v$ be a node in a graph, the neighbors of $v$ is the set of nodes that
points to (or, \emph{passes messages} to) $v$, formally:
 $$
 NBR(v)=\{i|(i,v)\in E\}
 $$
 In the next subsection, we will discuss how $NBR(v)$ \emph{passes message} to $v$.
 But now, we need to lift the definition of $NBR$ to hypergraph.
We lift the idea of neighbors into left/right neighbors, which means each node can get message from both side in a hyperedge.
Formally:
\begin{align*}
  NBR_H(v)=&\{(l,r)|(l,v,r)\in \hat{E}\} \cup\\
                &\{(l,r)|(r,l,v)\in \hat{E}\} \cup\\
                 &\{(l,r)|(v,r,l)\in \hat{E}\}\\
  NBR_L(v)=&\{l|(l,r)\in NBR_H(v)\}\\
  NBR_R(v)=&\{r|(l,r)\in NBR_H(v)\}
\end{align*}
The task of machine learning on graph domain can be either
 \emph{graph-level} or \emph{node-level} .
In graph-level, a graph $G$ is mapped to a vector of reals
$\tau(G)\in\real^m$.
In node-level, $\tau$ depends on a node $n$ of $G$, i.e.
$\tau(G,n)\in\real^m$.
For example, compute the  size of graph is a graph-level task,
compute the degree of vertex is a node-level task.

\subsection{The Message Passing Framework}
\label{sec:mpnn}

The Message Passing Neural Network(MPNN)~\cite{MPNN} is a general framework for supervised learning on graphs.
It is originally a graph-level prediction framework for  chemical compound, we slightly modify it into node-level prediction.
The main idea of message passing  is to
embed each node into vector space, then iteratively refine the embeding.
In an iteration, \emph{each node receives messages from its neighbors and updates its embedding accordingly.}
In this paper, we also call embedding of node as  \emph{state}.

Let $h_v^t$ be the embeding of node $v$ at time $t$, $E_{iv}$ be the embeding of edge $(i,v)$,
and $a_v$ be a handcrafted feature of $v$. The $h_v^0$ is initialized by the zero-padding of $a_v$.
Formally, message passing  is defined by  message function $\mM_t$  and vertex update function $\mU_t$.
\begin{align*}
m_v^{t+1}&=\lambda*\sum_{i\in NBR(v)}{\mM_t(h_i^t,E_{iv},h_v^t)}\\
h_v^{t+1}&=\mU_t(h_v^t,m_v^{t+1})
\end{align*}
where
$\mM_t(h_i^t,E_{iv},h_v^t)$ is the message $i$ sent to $v$,
The $\lambda$ can be $1$ or ${1}/{|NBR(v)|}$ for different message aggregation 	strategies.
All the incoming messages of $v$ will be aggregated by average if $\lambda={1}/{|NBR(v)|}$.
For $\lambda=1$, the messages is aggregated by suming up.
After the message passing, we \emph{read out} each prediction $y_v$ of node $v$,
 from its final refined embeding $h_v^T$ and handcrafted feature $a_v$ 
$$
y_v=\mO(h_v^T,a_v)
$$
We collect  the prediction of all node $y=(y_1, y_2, ..., y_{|V|})$ as the output of neural network.

Notice that the $\mM,\mU,\mO$ are all \emph{undefined} by now.
While the MPNN is a framework. Each design of $\mM,\mU,\mO$ defines a concrete Neural Network.
For example, Gated Graph Neural Networks (GGNN)~\cite{GGNN},  Deep Tensor Neural Networks
(DTNN)~\cite{DTNN} are all instance of MPNN, which define two different $\mM,\mU,\mO$.
In fact, The MPNN originally came from the abstraction of at least eight notable NNs that operate on graphs.
Our work of lifting  message passing is on the message function $\mM_t$.
For implementation, we will use GGNN as the instance of MNPP in this paper.

\subsection{Gated Graph Neural Network}
\begin{figure}[htbp]
  \centering
  \subfigure[Graph]{
    \includegraphics[width=0.19\textwidth]{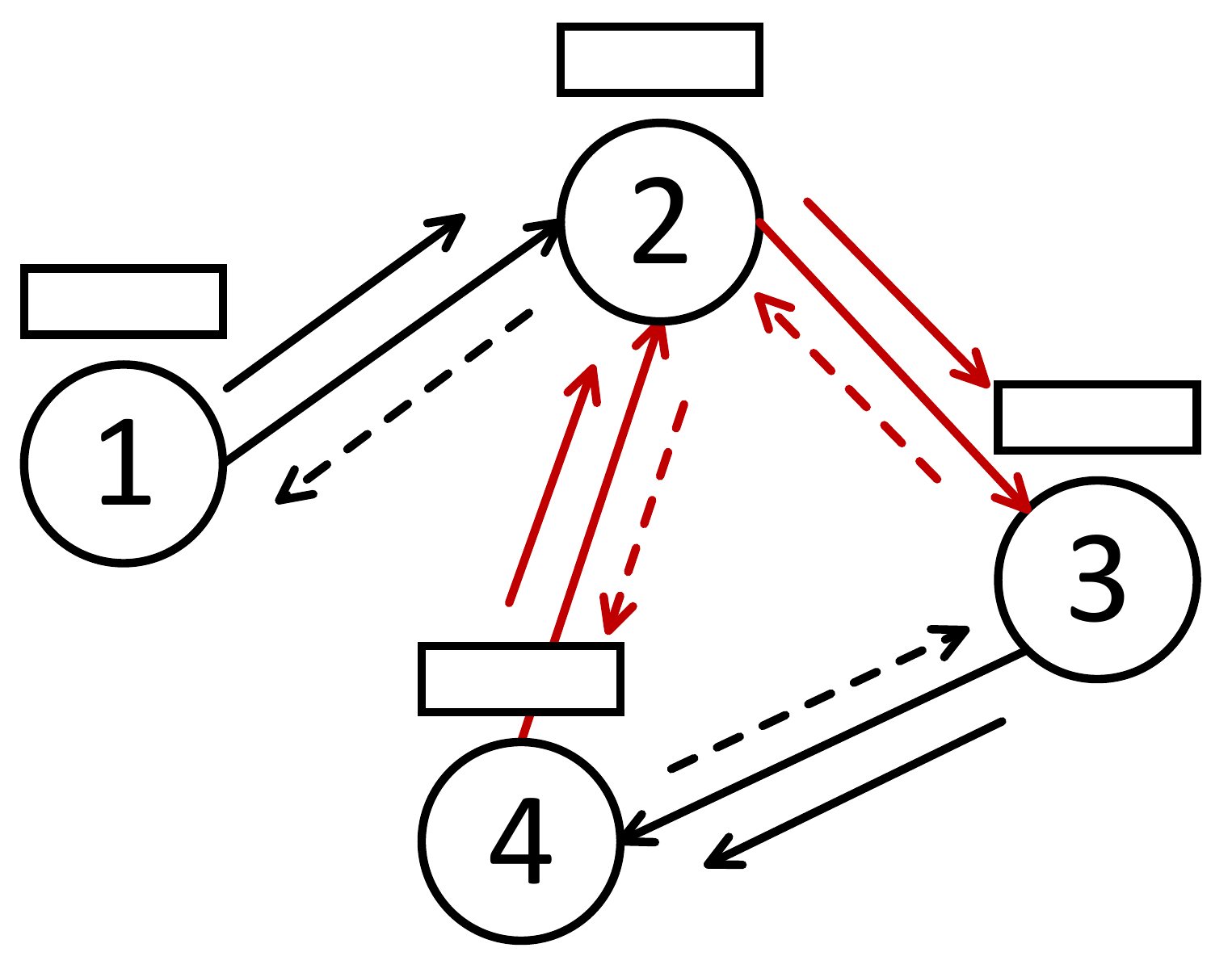}
    \label{fig:result16}
  }
\hspace{0.001\textwidth}
  \subfigure[Message Passing]{
    \includegraphics[width=0.25\textwidth]{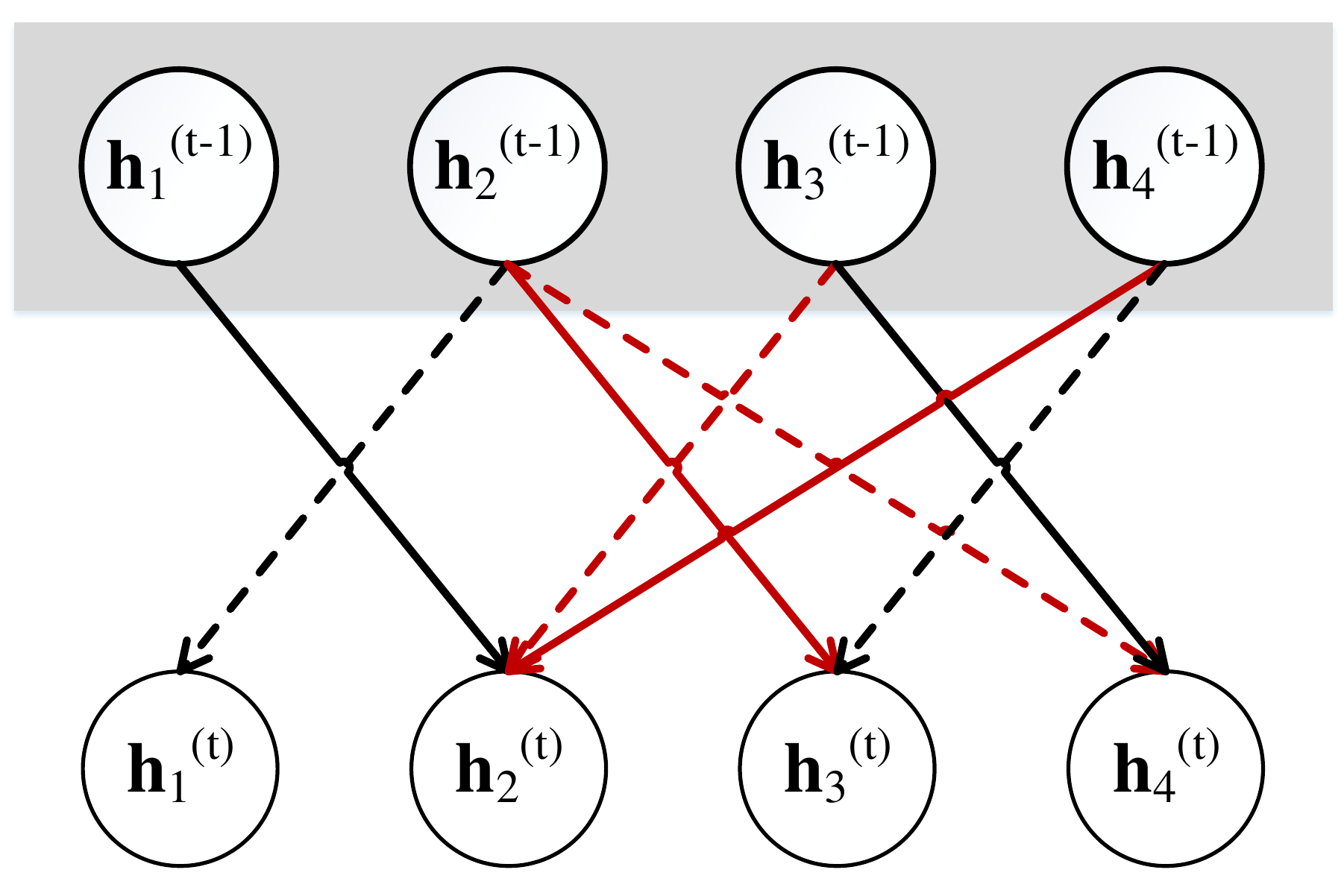}
    \label{fig:result30}
  }
  \caption{Example of Gated Graph Neural Network}
  \label{fig:result}
\end{figure}

GGNN assumes the labels on edges are finite and discrete, and calls the labels \emph{types}.
Let's take  molecules as examples. We saw atoms as vertices,
the \emph{type} of edges can be the chemical bond (single bond, $\pi$ bond, etc).
However, the distance of atoms cannot be the edge type, since distance is not discrete.

In GGNN, the embeding of node is in vector space $h_v^t \in \real^h$,
the embeding of edge is in matrix space $E_{i,j}\in \real^{h\times h}$.
We use  matrix $A_k$ to model the $k^{th}$ type of edges.
The parameter of $A_k$ is a  learned from training of neural network.
Let $\gamma$ be the type of edge $E_{i,j}$,
the embeding of edge of determined by its type $E_{i,j}=A_\gamma$


The  message function is designed as \emph{matrix multiplication} $\mM_t(h_i^t,E_{iv},h_v^t)=E_{iv}h_i^t$.
The update function is $\mU_t(h_v^t,m_v^{t+1})=GRU(h_v^t,m_v^t)$, where GRU is the Gated Recurrent Unit  introduced in~\cite{GRU}.
The same update function is used at each time step t.
Finally $\mO(h_v^T,x_v)=g_\theta([h_v^T,x_v])$, where $g_\theta$ is a fully connected neural network, and $[\cdot,\cdot]$ means concatenation of vectors

Several GGNN can by composed successively as several layers \cite{GGNN} in a way that
the output (i.e. final state) of the current message passing process
is used as the initial state of the next message passing process.
In each layer (i.e. message passing process), the message passing is repeated
for several times, with the same parameters of NN.
But different layers have different parameters.
We denote $h_v^{(k,t)}$ as the state of node $v$ on the $t^{th}$ timestep in
the $k^{th}$ layer,
and $T_k$ as the number of timesteps in the $k^{th}$ layer.
The layered GGNN can be formalized as $h_v^{(k+1,0)}=h_v^{(k,T_k)}$.

The idea of residual connection (i.e. skipping over layers)~\cite{resNet} can also be introduced into the connection of GGNN layers.
The incoming message of each node can be concatenated to the final state of
\emph{several previous layer} before that is fed into $\mU$.
For example, the message of each node $v$ in the $4^{th}$ GGNN layer, can be
concatenated to the final state of $0^{th}$ and $2^{rd}$ layers.
$$
h_v^{(4,t+1)}=\mU_{(4,t)}(h_v^{(4,t)},[m_v^{(4,t+1)},h_v^{(0,T_0)},h_v^{(2,T_2)}])
$$
The residual connection is used to reduce the problem of vanishing gradients in backpropagation.

  \section{OBDD Reordering as DL on Hypergraph}

Neural network (NN) has been proven a powerful machine learning technique for nonlinear data-fitting
problem~\cite{hornik1989multilayer}.
In this section, we show how an OBDD problem can be reduced to a deep learning
problem on 3-uniform hypergraph.
We utilize NN to learn the patterns of ``good'' OBDD variable orders from real-world example.
After the training phase, NN can predict a good variable order for a given 3-CNF formula in a short time.

\subsection{Inputs}

The input of neural network is a 3-Hypergraph.
The labels on hyperedges are  finite and discrete, we call it \emph{types} just like what we did in normal graph.
Each 3-CNF is converted to a 3-Hypergraph.
Let $X=\{x_1,x_2,...,x_N\}$ be the variable set of a given CNF.
The vertex set $V$ of the converted hypergraph $H=(V,\hat{E})$ is $V=X\cup\{x_\perp\}$,
where the $x_\perp$ is a special node that represents $False$.
Each clause in 3-CNF is converted to a hyperedge directly from it's variables.
The type of each hyperedge is decided by the type of each literal(i.e $+$ and $-$).
Especially, the type of literal $False$ is $0$.
 For example:
the clause $x_{325}\lor \bar{x}_{174} \lor x_{299}$ is converted to the
hyperedge $(x_{325},x_{174},x_{299})$ with the type $+-+$,
and the clause $\bar{x}_5 \lor x_7$ is converted to the hyperedge
$(x_5,x_7,x_\perp)$ with the type $-+0$.
For simplicity, we use $(a,b,c)$ to represent the hyperedge $(x_a,x_b,x_c)$.

To start the message passing, each node $v$ needs a handcrafted feature $a_v$ to initialize  $h_v^0$.
We sort those variables primarily by the frequencies of occurrence, secondarily by the frequency of positive literal if variables appear same times.
Lastly we use lexicographic order of variable name if they are still same.
We use the position of $v$ in the sorting order
to construct an one-hot vector as $a_v$.
If $v$ is the $i^{th}$ variable, the $i^{th}$ element of $a_v$ is 1, other elements are 0s
For $a_\perp$ we use zero vector to initialize.
Let us take $(x_1\lor x_2)(\bar{x}_1\lor x_2 \lor x_3)$ as an example,
we use
\begin{align*}
  a_\perp=(0,0,0)~
  a_1=(0,1,0)~
  a_2=(1,0,0)~
  a_3=(0,0,1)
\end{align*}
as handcrafted features, use zero-padding $h_v^0=(a_v,0,\cdots)\in \real^h$ to initialize $h_v^0$.
This encoding method ensures $h_v$ almost independent of  the    name of Boolean variable $v$.

It should be noted that, the hypergraph is only converted from CNF,
which is independent from its graph of BDD.
The graph of OBDD \emph{has no relation} with NN in this paper.
%
%
\subsection{Outputs}
\label{sec:ddepth}
Outputs of the OBDD reordering problem are variable orders. 
We want the neural network to find a near-optimal order in a short time.

A variable order can be specified as a permutation of variables.
For example, the variable orders of the two OBDDs
in~\Cref{fig:ObddPic} are $x_0 x_1 x_2 x_3 x_4 x_5$ and
$x_0 x_2 x_4 x_1 x_3 x_5$, respectively.
However, the variable permutation is not a proper format of the neural
network's output.
Generally, a neural network requires its output to be a differentiable structure such
that the gradient descent algorithm can work on~\cite{rumelhart1986learning}.

To this end, we let the output of the OBDD reordering problem to be a vector of
real numbers, called the \emph{depth vector}.
Formally, given a variable $x\in X$, denote $\depth(x)\in \mathbb{R}$,
\zap{which is a real number
between 0 and 1, the differentiable depth of $x$.}
For example, a depth vector of the right OBDD in \Cref{fig:ObddPic}
is $\langle 1.7, 6.7, 3.3, 8.3, 5.0, 1.0\rangle$.
The less the depth value is, the more front the corresponding variable
in the order.
With the above depth vector, we
can quickly decide the variable order: $x_0\prec x_2\prec x_4\prec x_1\prec x_3\prec x_5$.

\subsection{Loss Function}
After the definition of input and output, we also need a suitable loss function for out task.
Since the final order computed from the final depth vector
is only related to the order of the values, but not the detailed values in the vector.
We use the angle $\theta$ of the predicted vector to the expected vector to measure the error, i.e., 
$$
loss(y,y^*)=\theta(y,y^*)=arccos(\frac{y\cdot y^*}{ \lVert y \rVert \lVert y^* \rVert})*\frac{180^\circ}{\pi}
$$
where $y$ is the prediction of NN, $y^*$ is the target vector, and $\lVert \cdot \rVert$ means 2-norm.
Notice that, each element $y^*$ is the near-optimal OBDD depth of corresponding variable (i.e. node in hypergraph).
We have already convert the OBDD reordering task into node-level learning task on hypergraph.
We don't care the state of whole hypergraph but do focus on the prediction of each node (i.e. depth of variable).

  \section{Neural Network for 3-Hypergraph}
In this section, we discuss how to generate  messages on hyperedges.
Firstly we define a message function $\hat{\mM}$ on hyperedges,
then introduce non-square matrices to model each hyperedge.
We also find a method to convert one hypergraph $H=(V,E)$ to two ordinary
graphs, $G_H=(V,E)$ and $G_H^{-1}=(V,E^{-1})$.
The incoming message of $v\in V$ in $H$ equals the sum of messages in $G_H$ and $G_H^{-1}$.
The \NNsname~can thus be implemented on the top of MPNN.


\subsection{Lifting Message Passing to HyperEdge}
As we already discussed in \Cref{sec:mpnn},
the message functions $\mM_t$ are used to generate message on edges.
Messages are used to update the states of vertices in graph, to learn the embeding of each type of edge from massive data.
Following this idea, the first thing we need to do, is to design a form of message function which can generate message on 3-hyperedges.
To achieve this goal, message function $\hat{\mM}$ should be defined on hyperedge:
\begin{small}\begin{align*}
\hat{m}_v^{t+1}&=\hat{\lambda}*\sum_{(i,j)\in NBR_H(v)}{\hat{\mM}_t(h_i^t,\hat{E}_{ivj},h_j^t,h_v^t)}
\end{align*}\end{small}
where $\hat{\lambda}$ can be either $1$ or $1/|NBR_H(v)|$. What needs to be emphasized is that,  the  modification of message function is the only modification of the MPNN framework.
The update function $\mU$ and the readout function $\mO$ all remain the same.

Our motivation is to keep the  framework of message pass unchanged, but lift the message generation on hyperedges.

\subsection{Hyperedge Message Functions}
We have already extended the framework of MPNN into hypergraph,
now we discuss how to implement the message functions in GGNN.
The key idea of GGNN is to use square matrices to model ordinary edges.
Each edge type is modeled by a matrix $A_k \in \real^{h \times h}$.
Finally GGNN uses a matrix multiplication $E_{iv}h^t_i$  to implement the message function and generate messages.
The $A_k$ can be seen as a mapping from node state to message: $A_k: \real^h \rightarrow \real^h$.
We need to lift the mapping into $\hat{A}_k:\real^{2h} \rightarrow \real^h$ since one node gets two neighbors in a hyperedge.
So we  lift the square matrix $A_k \in \real^{h \times h}$ into non-square matrix $\hat{A}_k \in \real^{h \times 2h}$,
and use $\hat{E}_{ivj}\colvec{h^t_i}{h^t_j}$ to implement message functions:
$$
\hat{\mM}_t(h_i^t,\hat{E}_{ivj},h_j^t,h_v^t)=\hat{E}_{ivj}\colvec{h^t_i}{h^t_j}\in \real^h
$$
Now we have a method to generate and pass messages on hypergraphs.
We also find a way to implement the \NNsname~on the top of existing MPNN.

\subsection{Implementation on The Top of MPNN}

Since the matrix can be partitioned into  blocks.
The hyperedge message function $\hat{\mM}$ can also be \emph{rewritten} as block matrix multiplication.
Following this idea, We surprisingly found that the $\hat{\mM}$ can be reduced to
two \emph{ordinary message passings} on ordinary graph.
This makes it possible to implement \NNsname~on the top of the existing MPNN.
The key is to decompose the hyper message passing.
Firstly we divide the matrix into 2 blocks, i.e. $\hat{E}_{ivj}=[E_{iv},E_{vj}]$, then:
\begin{small}
\begin{align*}
  \hat{m}_v^{t+1}&=\hat{\lambda}*\sum_{(i,j)\in NBR_H(v)}\hat{\mM}_t(h_i^t,\hat{E}_{ivj},h_j^t,h_v^t) \\
      &=\hat{\lambda}*\sum_{(i,j)\in NBR_H(v)} \hat{E}_{ivj}\colvec{h^t_i}{h^t_j}\\
        &=\hat{\lambda}*\sum_{(i,j)\in NBR_H(v)}[E_{iv},E_{vj}]\colvec{h^t_i}{h^t_j}\\
        &=\hat{\lambda}*\sum_{i\in NBR_L(v)}E_{iv}{h^t_i}+\hat{\lambda}*\sum_{j\in NBR_R(v)}E_{vj}{h^t_j}
\end{align*}
\end{small}
We call a graph $G_H=(V,E)$ a derived graph of $H=(V,\hat{E})$ when
\begin{small}
\begin{align*}
E=\{(i,j)|(i,j,k)\in \hat{E}\} &\cup \{(j,k)|(i,j,k)\in \hat{E}\} \\
&\cup \{(k,i)|(i,j,k)\in \hat{E}\}
\end{align*}
\end{small}
and denote $G_H^{-1}=(V,E^{-1})$ as the reverse graph of $G_H$ where $E^{-1}=\{(j,i)|(i,j)\in E\}$.
We get that
\begin{small}
\begin{align*}
  \hat{\lambda}*\sum_{i\in NBR_L(v)}E_{iv}{h^t_i}&=\overrightarrow{m}_v^{t+1}\\
  \hat{\lambda}*\sum_{j\in NBR_R(v)}E_{vj}{h^t_j}&=\overleftarrow{m}_v^{t+1}
\end{align*}
\end{small}
where $\overrightarrow{m}_v^{t+1}$ is the message of $v$ on $G_H$, $\overleftarrow{m}_v^{t+1}$ is the message of $v$ on $G_H^{-1}$.
Finally we get
$$
\hat{m}_v^{t+1}=\overrightarrow{m}_v^{t+1}+\overleftarrow{m}_v^{t+1}
$$
which means that the MPNN can be used to implement the message passing of
hypergraph by decomposing a hypergraph into  the derived graph and it's reverse.

  \section{Implementation and Evaluation}
We implement our algorithm on the top of Tensorflow~\cite{abadi:tensorflow},
and used ADAM~\cite{kingma:adam} for the learning rate control.
All experiments 
were performed on GeForce GTX 1080 Ti GPU and an Intel Xeon E5 CPU.

To evaluate the efficiency of our approach, up to 7 typical OBDD reordering
algorithms are compared:

\begin{itemize}
    \item \NNsname: our neural network approach;
    \item GA: the genetic algorithm~\cite{drechsler1996genetic} for OBDD reordering.
    \item LINEAR: a combination of sifting variable up and down and linear transformations of two adjacent variables\cite{linear}.
    %
    \item RAND: randomly choose pairs of variables and swap them in the
        order~\cite{somenzi2015cudd}.

    \item G-SIFT: the group sifting method~\cite{panda1995variables}.
    \item WIN2: the iterating window algorithm~\cite{felt1993dynamic}, with its window size
        been set to 2. 
    \item WIN3: also the~\cite{felt1993dynamic}, but window size is 3.
\end{itemize}
In \NNsname, we  embed each node to 500-dimensional vector space.
We have $5$  GGNN layers, the each layer correspondingly propagate $2, 2, 1, 2, 1$ times.
The $2^{rd}$ layer has residual connections from $0^{th}$ layer, and
the $4^{rd}$ layer has residual connections from both $0^{th}$ and $2^{th}$ layer.
We use average function to do message aggregation.

%

\subsection{Data Set and Benchmark}
We choose the LGSynth91 benchmark~\cite{yang1991logic} as our data set.
We collect the circuits in Berkeley Logic Interchange Format (blif)~\cite{blif} format, convert them to And-Inverter Graph(aig)~\cite{aig} format, extract the transition relation boolean formula into equisatisfiable 3-CNF.
Note that the genetic algorithm~\cite{drechsler1996genetic} attains the best result among all
OBDD reordering algorithms.
For each sample, we run the genetic algorithm to compute the
near-optimal variable order, and use this order as the label.
We set a time-out of 30 minutes, with the time of building the initial BDD and
the reordering being all counted.
There are 28 samples that can finish GA labeling in 30 minutes.

While it's far not enough to train a Neural Network,
so we  randomly mutate the circuit in the level of aig: randomly negate a variable of an and-gate 1$\sim$3 times.
For each sample we make 200 distinct mutations and then run GA labeling again on them.
Finally we get 5138 labeled samples, among which 80\% are used for training
and the rest are used for testing.
The evaluation is performed \emph{only} on the test set.
The  number of variable and clause varies from $11\sim220$ and $18\sim633$.
The phase(i.e. clause/variable) varies from $1.6\sim2.9$.
The average numbers of variable, clause, phase are 59.3, 139.3, 2.3 correspondingly.
After the training, the best loss we can get on test set is $\theta=20.6^\circ$.

We also take a step forward, make  a more challenging stress test on our \NNsname.
We collect some samples of LGSynth91 (with less than 300 variables) that can not finish the GA labeling in
time limit, and call them the \emph{hard benchmark}.
The samples in the hard benchmark are all challenging enough for OBDD.
We are very curious about the performance of \NNsname~on the hard benchmark.


\subsection{Results on Time}
To evaluate the efficiency of those algorithm, we compare their computation time of giving a result of near-optimal order.
We only consider the time of perform algorithms, the time of building initial OBDD is not included.
%
The result of average computation time (seconds) is in \Cref{table:AverageComputationTime}.

  \begin{table}[htbp]
    \scalebox{0.72}{
\begin{tabular}{@{}cccccccc@{}}
\toprule
Algorithm & GA & LINEAR & \NNsname & G-SIFT & RAND & WIN3 & WIN2 \\ \midrule
Time(sec) & 43.50 & 12.29 & \textbf{0.01} & 12.92 & 9.65 & 0.58 & 0.24 \\ \bottomrule
\end{tabular}}
\caption{Average Computation Time}
\label{table:AverageComputationTime}
\end{table}

\begin{figure}[htbp]
  \centering
  \includegraphics[width=0.45\textwidth]{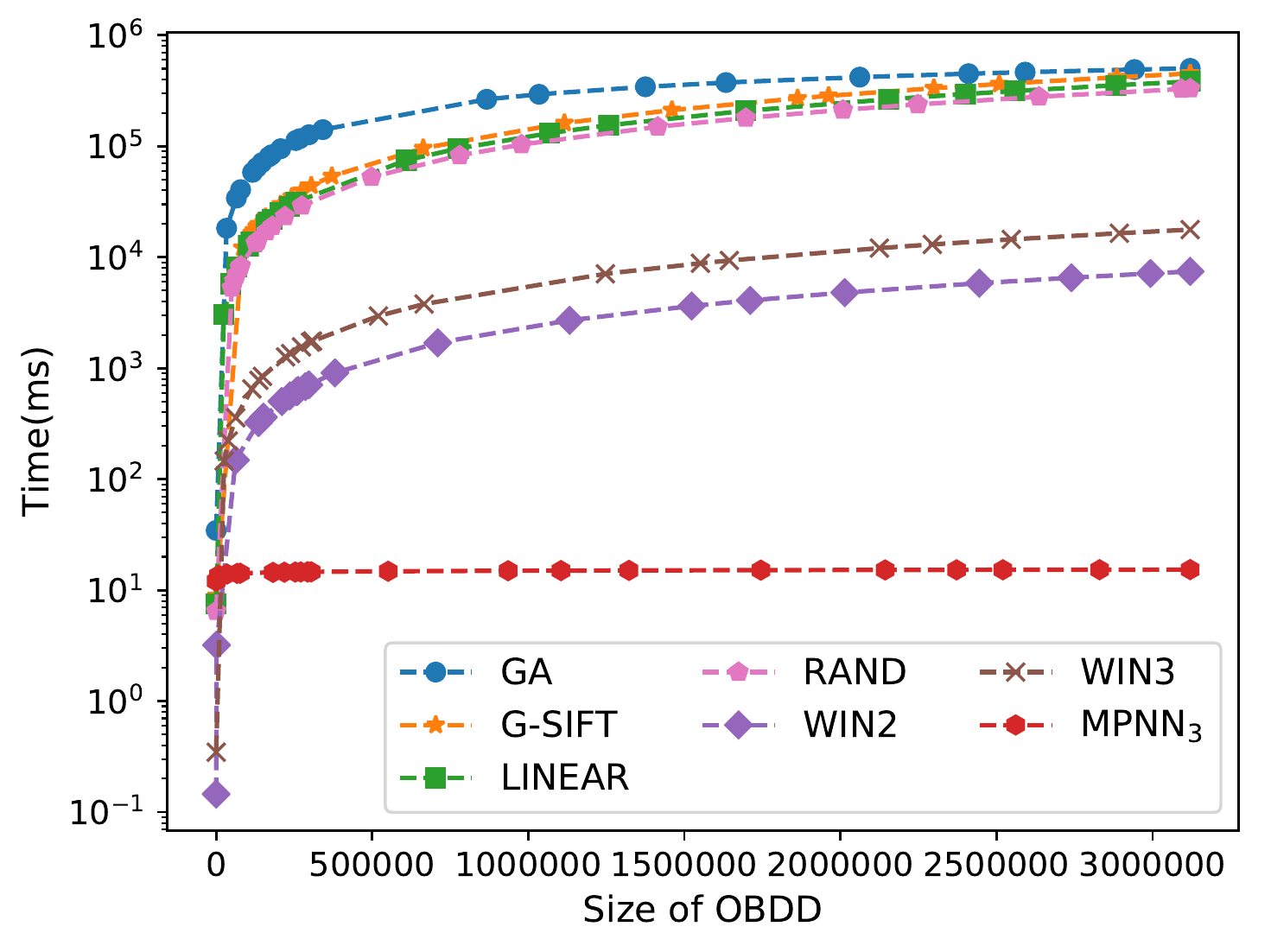}
    \caption{Curve of OBBD Size and Computation Time}
    \label{fig:CurveTime}
\end{figure}

The WIN2 and WIN3 are quite efficient among the traditional methods.
The GA takes longest time to give a best result.
The RAND makes balance between compression ratio and time.
However, \NNsname~is the fastest algorithm.
We go further and    fit a curve of time for each algorithm in \Cref{fig:CurveTime}.
The horizontal axis lists the sizes of the input OBDDs,
the vertical axis shows the average computation time of different reordering algorithms.
Note that the vertical axis is logarithmic.

We observe that GA slows down quickly with the increasing of OBDD’s size.
In contrast, \NNsname~is not sensitive to the size of the input OBDD.
Recall that the inputs of the \NNsname~are CNFs, instead of OBDDs.
To conclude, our approach can get a near-optimal variable order in a short time.
But will such a fast speed of \NNsname~affect the quality of its solution?

\subsection{Results on Compression Ratio}
To evaluate the accuracy of reordering algorithms,
we compare their compression ratios.

Given a Boolean function $f$ and a variable order, we denote the
original OBDD by $G$. All reordering algorithms are respectively applied to
$f$ to produce a new variable
order. The OBDD with respect to the new order is denoted by $G'$.
In the experiments, we use CUDD~\cite{somenzi2015cudd} to evaluate the corresponding OBDD's size.
If $|G'| < |G|$, we adopt the new order.
Let $A$ be a reordering algorithm, we use
    $\eta_A = \left( {|G'|-|G|} \right)/ {|G|}$
to measure the compression ratio.
\begin{figure}[htbp]
  \centering
  \includegraphics[width=0.47\textwidth]{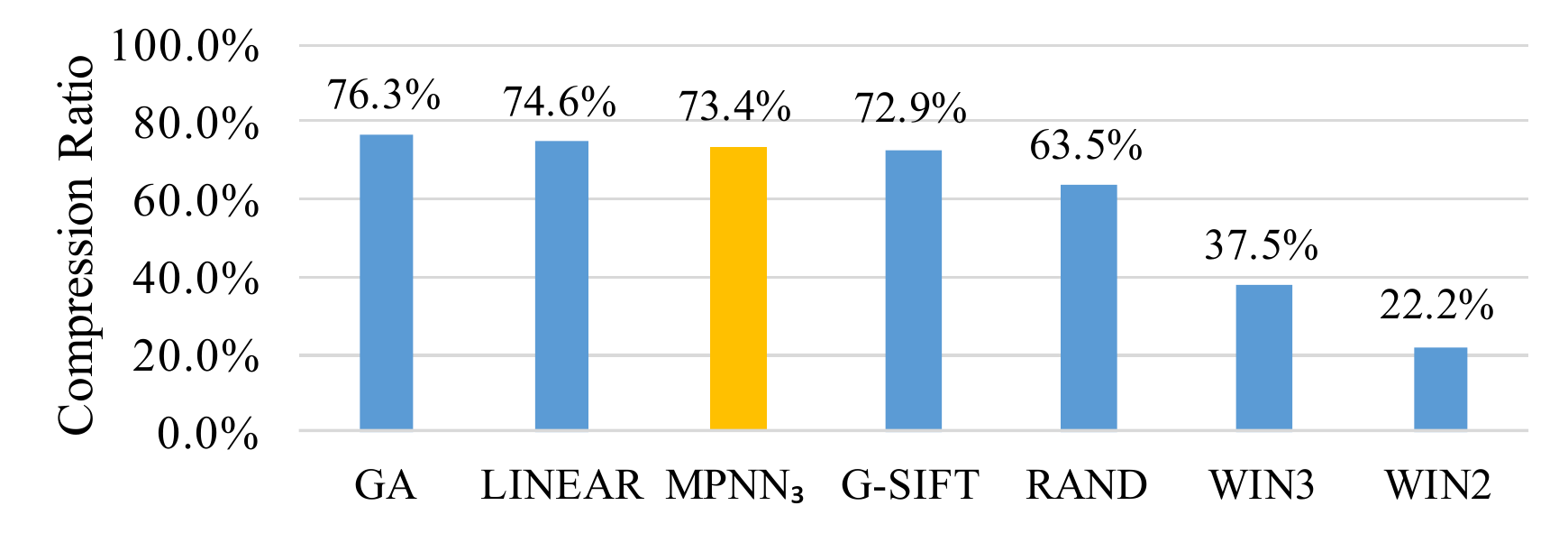}
  \caption{Average Compression Ratio}
  \label{fig:AverageCompressionRatioBars}
\end{figure}
The average compression ratios of each algorithm on test set are shown in \Cref{fig:AverageCompressionRatioBars}
The horizontal axis indicates 7 algorithms and the vertical axis shows their compression ratios.
From \Cref{fig:AverageCompressionRatioBars}, observe that GA always gets the best compression ratio. This conforms to the existing results in literatures~\cite{drechsler1996genetic,jindal2017novel}.
The \NNsname~achieves $3^{rd}$ best result,
and the results of the top-4 algorithms are close.
\begin{figure}[htbp]
  \centering
  \includegraphics[width=0.45\textwidth]{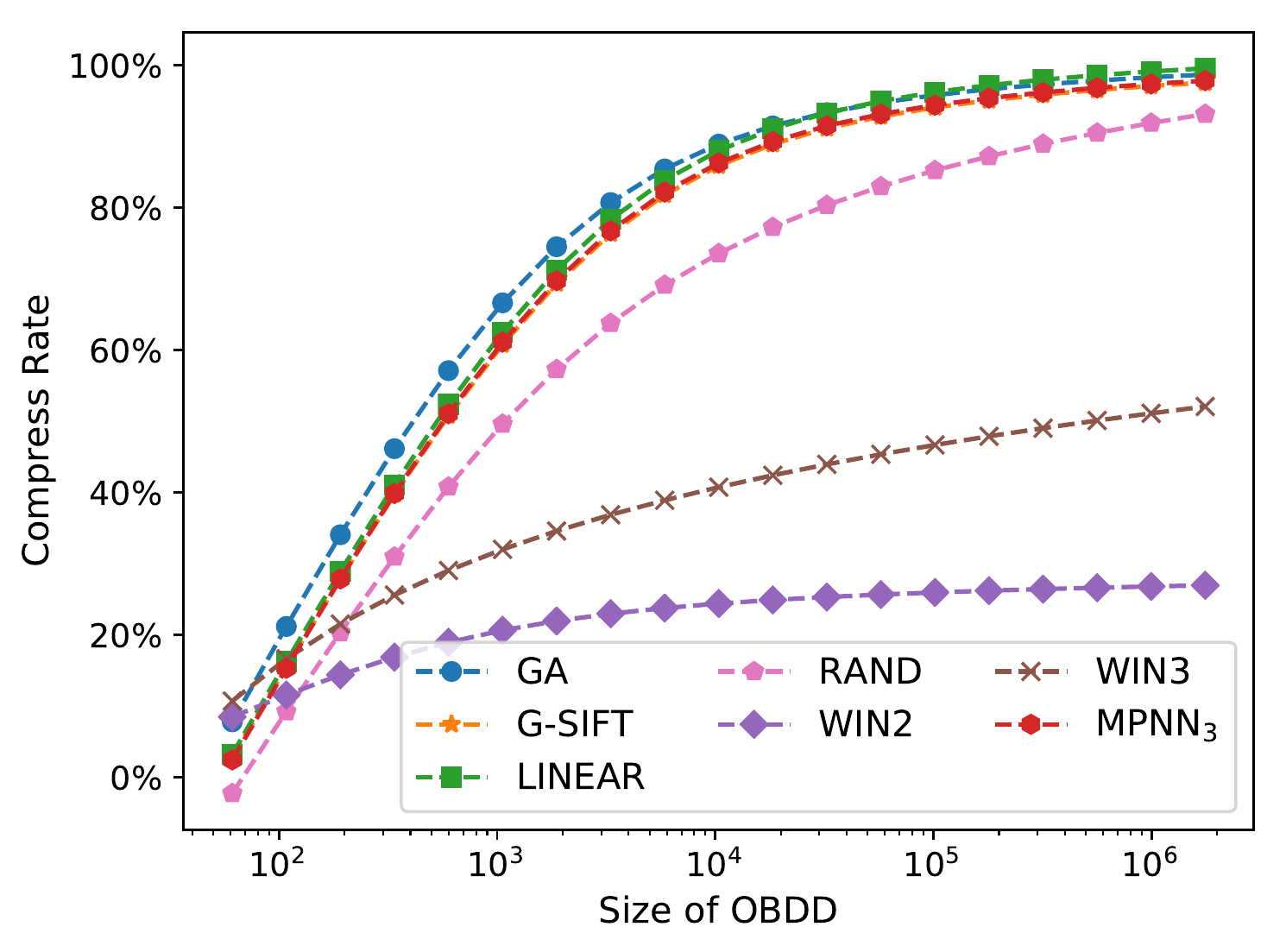}
  \caption{Curve of OBBD Size and Compression Ratio}
  \label{fig:CurveCrate}
\end{figure}
To see more details of those sample, we fit a curve of compression ratio for each algorithm in \Cref{fig:CurveCrate}.
The horizontal axis lists the sizes of the input OBDDs,
and the vertical axis shows the average compression ratio of different reordering algorithms.
 Note that the horizontal axis is logarithmic.
 We find that smaller  OBDDs are harder  to be compressed.
 This is understandable, since the difference between linear and exponential OBDD size is smaller when the number of variable is smaller.
The curves of top-4 algorithms are close and GA is always better then other algorithms.
The WIN2 and WIN3 is not such effective but quick.
How \NNsname~performs in hard benchmark? we will talk it in next subsection.

\subsection{Results on Stress Test}
It is  challenging for BDD-method in large circuits.
We set the timeout for 12 hours, and give 110GB memory for each samples.
Firstly, there are 46.2\% of hard benchmark cannot even build an initial OBDD,
  we call them \emph{very-hard benchmark} for simplicity.
50\% of very-hard benchmark are out of time for 12 hours (OOT), others are out of memory for 110GB (OOM).
The traditional methods are performed on the initial OBDD, so they are failed on those task.
However, recall that the prediction of \NNsname~doesn't rely on the initial OBDD.
We directly use the order of \NNsname~to build OBDD.
41.7\% samples in very-hard benchmark can build the OBDD using the order of \NNsname,
others are all OOT, not OOM, which means it still has some possibility for them to build OBDDs if we give more time.

For the rest of hard benchmark, which traditional method can be performed,
we compare \NNsname~with Win2, Win3, Rand. 
There are 2 samples OOT for Rand, we lists some results in \Cref{table:ResultOnHard}


\begin{table}[htbp]
  \centering
  \tabcolsep2.2pt
  \scalebox{0.84}{
\begin{tabular}{@{}llr|rr|rr|rr|rr@{}}
  \toprule
  Name & Vars & Nodes & \multicolumn{2}{c|}{\NNsname} & \multicolumn{2}{c|}{WIN2} & \multicolumn{2}{c|}{WIN3} & \multicolumn{2}{c}{RAND} \\ \midrule
  cordic & 106 & 9M & 99\% & 0.01 & 24\% & 17 & 47\% & 50 & 96\% & 1701 \\
  s298 & 133 & 2M & 80\% & 0.01 & 42\% & 1 & 67\% & 4 & 93\% & 122 \\
  s344 & 144 & 41M & 97\% & 0.01 & 24\% & 71 & 34\% & 271 & 98\% & 11194 \\
  s349 & 148 & 47M & 90\% & 0.01 & 25\% & 73 & 36\% & 279 & 99\% & 6658 \\
  mux & 153 & 10M & 85\% & 0.01 & 30\% & 9 & 47\% & 37 & 99\% & 537 \\
  sct & 159 & 3M & 83\% & 0.01 & 17\% & 2 & 53\% & 6 & 96\% & 61 \\
  lal & 164 & 219M & 99.7\% & 0.01 & 14\% & 444 & 47\% & 1807 & \multicolumn{1}{c}{-} & OOT \\
  s382 & 185 & 12M & 52\% & 0.01 & 44\% & 13 & 64\% & 31 & 91\% & 1232 \\
  s386 & 185 & 0.5M & 40\% & 0.01 & 12\% & 0.4 & 20\% & 1 & 75\% & 17 \\
  s400 & 193 & 13M & 62\% & 0.01 & 43\% & 12 & 63\% & 30 & 92\% & 2580 \\
  s444 & 193 & 11M & 89\% & 0.01 & 16\% & 17 & 18\% & 58 & 97\% & 1123 \\
  s420 & 210 & 43M & 93\% & 0.01 & 27\% & 69 & 43\% & 197 & 96\% & 5732 \\
  s510 & 244 & 7M & 99\% & 0.01 & 38\% & 5 & 41\% & 22 & 99\% & 438 \\
  s526 & 248 & 594M & 92\% & 0.01 & 10\% & 1433 & 52\% & 5251 & \multicolumn{1}{c}{-} & OOT\\
  \bottomrule
\end{tabular}}
\caption{Result On Hard Benchmark}
\label{table:ResultOnHard}
\end{table}

The first column is the name of samples,
the second column is the number of variables.
The third column is the size of initial OBDD, where the `M' means million($10^6$).
Others are result of each algorithm.
The first column of each algorithm result is the compression ratio, the second column is time in seconds.
The result shows that \NNsname~achieves a very good result in the stress test, totally beats WIN2 and beats WIN3 in most case.
The compression ratios of \NNsname~is also competitive to RAND, with 2 case can not finish measure using RAND in 12 hours.
The speed of \NNsname~is extremely fast.

  \section{Conclusions}

In this paper, we apply \NNsname~to minimize OBDDs,
lift the neural message passing on 3-hypergraph to recieve 3-CNF as input.
We perform experiments to compare our approach to classical algorithms on
variable reordering of OBDDs.
Experimental results show that our approach can get competitive compression ratio in
an extremely short time.
There are many complex relationships in real world can be modeled by hypergraphs.
In the future, we plan to apply  \NNsname~to more fields.

  \bibliographystyle{aaai}
  \bibliography{./sections/reference}
\end{document}